\documentclass[letterpaper, 10 pt, conference]{ieeeconf}  
\IEEEoverridecommandlockouts                              
\usepackage{hyperref}
\usepackage{booktabs}
\usepackage{multirow}
\usepackage{xcolor}
\usepackage[ruled,vlined]{algorithm2e}
\usepackage{fancyhdr,amssymb}
\usepackage{algpseudocode}
\usepackage{amsfonts}
\usepackage{amsmath}
\usepackage{graphicx}
\usepackage{textcomp}
\usepackage{balance}
\usepackage{subcaption}

\title{\LARGE \bf
Controllable Motion Synthesis and Reconstruction \\ with Autoregressive Diffusion Models
}

\author{
    Wenjie Yin$^{1}$, 
    Ruibo Tu$^{1}$, 
    Hang Yin$^{1}$, 
    Danica Kragic$^{1}$, 
    Hedvig Kjellström$^{1}$, 
    Mårten Björkman$^{1}$
    \thanks{
        $^{1}$Division of Robotics, Perception and Learning, KTH Royal Institute of Technology, Sweden.
        {\tt\small \{yinw, ruibo, hyin, dani, hedvig, celle\}@kth.se}.
    }
}

\begin{document}
\maketitle
\thispagestyle{empty}
\pagestyle{empty}

%%%%%%%%%%%%%%%%%%%%%%%%%%%%%%%%%%%%%%%%%%%%%%%%%%%%%%%%%%%%%%%%%%%%%%%%%%%%%%%%
\begin{abstract}

Data-driven and controllable human motion synthesis and prediction are active research areas with various applications in interactive media and social robotics. Challenges remain in these fields for generating diverse motions given past observations and dealing with imperfect poses. This paper introduces MoDiff, an autoregressive probabilistic diffusion model over motion sequences conditioned on control contexts of other modalities. Our model integrates a cross-modal Transformer encoder and a Transformer-based decoder, which are found effective in capturing temporal correlations in motion and control modalities. We also introduce a new data dropout method based on the diffusion forward process to provide richer data representations and robust generation. We demonstrate the superior performance of MoDiff in controllable motion synthesis for locomotion with respect to two baselines and show the benefits of diffusion data dropout for robust synthesis and reconstruction of high-fidelity motion close to recorded data.

\end{abstract}

%%%%%%%%%%%%%%%%%%%%%%%%%%%%%%%%%%%%%%%%%%%%%%%%%%%%%%%%%%%%%%%%%%%%%%%%%%%%%%%%
\section{INTRODUCTION}
\label{sec:intro}

Motion synthesis techniques play an important role in computer animation, video games, human-robot interaction~\cite{valle2021transflower}, etc. Recently, significant progress has been achieved in motion generation and reconstruction by utilizing deep generative models~\cite{yin2021graph}, which can be broadly divided into deterministic and probabilistic models. Deterministic models~\cite{fragkiadaki2015recurrent, butepage2017deep} frame the motion synthesis task as a regression problem in which the response and input have exact relationships, leading to stereotypical results with limited diversity. 
In contrast, probabilistic models fit probabilistic distributions to the data distribution~\cite{henter2020moglow, wang2019combining}, which capture a range of motion patterns and as such significantly improve the motion diversity and fidelity.

Despite recent advances in deep generative models, motion synthesis still remains challenging in a number of aspects. For example, capturing complex relations between body limbs and motion frames requires models that are less susceptible to failures such as mode collapse. Also, robust and coherent synthesis is desirable even when long-term generation is conditioned on imperfect data. The latter is particularly demanding when human skeletal data are extracted from noisy sensors or previously generated frames. Earlier work often assume perfect conditioning patterns~\cite{henter2020moglow} or manually defined graph structures~\cite{yin2021graph}, and thus cannot satisfactorily mitigate this issue.

\begin{figure}[t]
\centering
\includegraphics[width=0.48\textwidth]{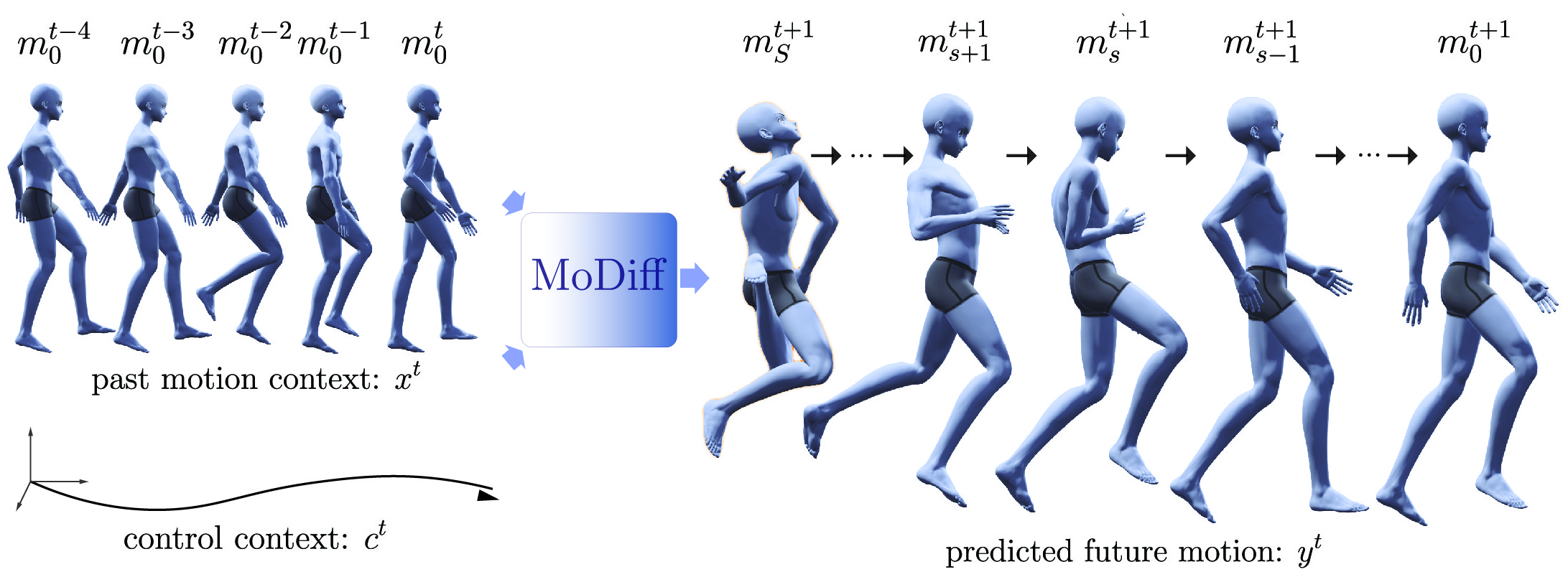}
\caption{ 
    Illustration of the generation process: MoDiff progressively denoises a noisy motion sequence to get a natural movement, given past motion and control contexts.  
  }
\label{fig:frontpage}
\end{figure}

In this paper, we propose MoDiff, a diffusion-based probabilistic model for high-quality controllable human motion synthesis, as illustrated in Fig.~\ref{fig:frontpage}.
Diffusion-based approaches have recently gained traction for their superior performance as probabilistic generative models and found application also in human motion synthesis~\cite{zhang2022motiondiffuse, tevet2022human}. 
Score-based diffusion models require no special neural network architectures, such as in the case of flow-based approaches~\cite{henter2020moglow, yin2021graph}, to optimize an exact likelihood. 
Our work leverages this flexibility by introducing a cross-modal Transformer-based architecture, which enables richer representations for encoding past motion frames and control contexts. The approach also exploits the intermediate representations generated in the diffusion process for a natural dropout strategy to improve robustness. 
Our evaluation on domain standard human locomotion datasets shows that MoDiff outperforms baseline methods and produces realistic motions conditioning on control contexts of other modalities. We further demonstrate that the same framework can be applied for reconstruction of imperfect motion sequences. 

In summary, our contributions are:
\begin{itemize}
\item We present a flexible neural architecture, MoDiff, that integrates multimodal transformer and autoregressive diffusion models for motion generation and reconstruction of missing parts in motion sequences. 
\item We propose diffusion data dropout, utilizing the forward process to obtain diffusion-induced motion representations, which can be employed to improve adherence to the control context. 
\item The proposed approach achieves superior results for human motion generation and reconstruction. Applications include, but is not limited to, locomotion synthesis, text-to-motion, and music-to-dance. 
\end{itemize}

The paper proceeds as follows: Section \ref{sec:related} gives an introduction to previous work on human motion synthesis, diffusion models, and data dropout. Section \ref{sec:method} formulates the problem and provides a comprehensive description of the proposed framework for controllable motion synthesis and reconstruction. Section \ref{sec:exp} introduces the experimental setup, discusses the results, and compares these to two baselines. Finally, Section \ref{sec:conclu} summarizes the paper and outlines future directions. 

\section{RELATED WORK}
\label{sec:related}

In this section, we provide an overview of deep learning-based human motion synthesis (Section \ref{sub:humanmotion}) and denoising diffusion probabilistic models (DDPMs) (Section \ref{sub: ddpms}), and then describe prior work on data dropout (Section \ref{sub: dropout}). 

\subsection{Human Motion Synthesis}
\label{sub:humanmotion}
Deep learning approaches have been widely adopted for human motion synthesis following earlier success in other domains. These approaches can be categorized into deterministic and probabilistic methods. Most previous works follow the deterministic approach to yield a fixed output for a given input. For instance, Fragkiadaki et al.~\cite{fragkiadaki2015recurrent} first applied recurrent neural networks to human motion prediction. Butepage et al.~\cite{butepage2017deep} directly fed the most recent previous frames through an encoder-decoder network to predict future motion frames. Li et al. \cite{li2017auto} introduced a conditioned LSTM for synthesizing long-term motion patterns, and Martinez et al. \cite{martinez2017human} employed a sequence-to-sequence architecture with residual connections for joint prediction. 

To synthesize motion patterns with more variety and diversity, probabilistic methods have also been adopted. Earlier works model distributions over human motion with Gaussian mixture models~\cite{crnkovic2016generative}, and Gaussian Process models~\cite{levine2012continuous}. 
As for deep neural networks, Variational Autoencoders (VAEs) and their variants, which optimize a lower bound on the data log-likelihood, have been used in combination with recurrent structures for controllable motion synthesis~\cite{habibie2017recurrent}. 
VAE-based approaches have also been utilized in cross-modal synthesis tasks, such as generating motion sequences from speech~\cite{greenwood2017predicting}.
Another significant branch of methods is based on Generative Adversarial Networks (GAN)~\cite{wang2019combining, yin2022dance}. GANs are deemed more powerful and effective for encoding, but are often difficult to train and evaluate. 
To this end, Flow-based generative models gained in popularity, as they enable tractable likelihood evaluation and efficient model parameterization. MoGlow~\cite{henter2020moglow} models motion sequences with autoregressive normalizing flow and recurrent neural networks. Yin et al.~\cite{yin2021graph} further integrated MoGlow with graph convolutional networks for motion reconstruction. 
Ho et al.~\cite{ho2020denoising} presented diffusion models, a new paradigm for probabilistic generative modeling that allows for greater flexibility in the choice of architectures. 
%Further reviews are provided in the following paragraph.
Our method adopts a diffusion model for improved model capacity and training stability in learning controllable motion synthesis. 

\subsection{Denoising Diffusion Probabilistic Models (DDPMs)}
\label{sub: ddpms}
DDPMs~\cite{ho2020denoising, song2019generative} are a class of generative models inspired by non-equilibrium thermodynamics. %sohl2015deep
DDPMs have a forward and a reverse process. The forward process progressively adds noise to data and the reverse process learns to construct data samples from the noise.  
The learning model allows regular neural networks to be used, which makes the models both analytically tractable and flexible. 

Several works on human motion generation have adopted diffusion models. %to generate human motion sequences. 
MID~\cite{gu2022stochastic} encodes historical information on behaviors and global signals as an embedding and devises a Transformer-based diffusion model for human trajectory prediction.
BelFusion~\cite{barquero2022belfusion} is a diffusion model that exploits a behavioral latent space for human motion prediction. 
The recent MotionDiffuse~\cite{zhang2022motiondiffuse} is considered to be the first diffusion model-based text-driven motion generation framework with instructions on body parts.
Similar to MotionDiffuse, MDM~\cite{tevet2022human} integrates diffusion models and CLIP~\cite{radford2021learning} for text-to-motion generation. 
Recently, diffusion models have been utilized to generate dance movements, a challenging task due to the intricate postures, rhythms, and compositions involved in dance, often accompanied by music. 
Alexanderson et al.~\cite{alexanderson2022listen} pioneer diffusion models with Conformer~\cite{zhang2022music} for audio-driven dance motion generation. 
Tseng et al.~\cite{tseng2022edge} propose EDGE, a transformer-based diffusion model that generates dance sequences conditioned on music. 
% MoFusion~\cite{ma2022pretrained} employs a Transformer backbone and diffusion models for unified motion synthesis. 

Our proposed framework applies diffusion models for motion generation and models the temporal information in an autoregressive manner, similar to TimeGrad~\cite{rasul2021autoregressive}, with a cross-modal transformer inspired by Li et al.~\cite{li2021ai} for controllable generation from various modalities. 
The difference is that our framework can impose/alter the control signal on the motion generation process on the fly while other models need a supplied command, whether it be text-based or in other forms, before the full sequence can be generated.
%in one shot and then wait until the completion of the whole motion. 
Moreover, the design of the autoregressive diffusion model is flexible and can be extended to tasks that require robust generation, e.g.~synthesis and reconstruction from imperfect motion frames, without additional training, something that is not featured in the works reviewed above. 

\subsection{Data Dropout}
\label{sub: dropout}
Due to an over-reliance on the autoregressive context, autoregressive models often suffer from poor adherence to the control context and as such have compromised consistency in controllable generation~\cite{henter2020moglow}. Such a phenomenon is exacerbated with long-term prediction. Natural approaches to counter this problem includes removing some or all of the conditioning information during learning. 
Bowman et al.~\cite{bowman2015generating} randomly replace some part of the conditioned word tokens with a generic unknown word token. They apply this technique to a decoder, helping the model to capture higher-order statistics. 
Wang et al.~\cite{wang2018autoregressive} propose a data dropout strategy that randomly sets the data to zero in both the training and generation stages, forcing the models to focus on the control context, thus alleviating this problem. 
Kovács et al.~\cite{kovacs2017increasing} propose an input channel dropout scheme, forcing the network to make decisions based on a subset of channels. 
To rememdy the poor adherence issue in the original  MoGlow~\cite{henter2020moglow}, MoGlow applies dropout to entire frames of the autoregressive past motion context. 

In this paper we proposes a new diffusion-induced dropout scheme. We leverage the Gaussian noise injected in the diffusion forward process and use intermediate representations as corrupted conditions. Our strategy naturally removes the requirement of tuning a separate process and is found to benefit the consistency between the generated motion and the control contexts. In addition, this also encourages encoders to learn richer and more robust representations. 

\section{METHODOLOGY}
\label{sec:method}

This section formulates our target problem and establishes notations used throughout the paper. Preliminaries about denoising diffusion probabilistic models are also given, including the training and inference strategies. On the basis of these, we introduce our contributed framework. 

\subsection{Problem Formulation}
\label{sub:formulation}
In our scenario, we treat the human motion sequence as a series of poses, and the aim is to synthesize the future motion and reconstruct the past imperfect poses using an autoregressive diffusion model.
Formally, a 3D skeleton-based pose at time step $t$ is denoted as $m^t$, with corresponding additional information $a^t$, such as control signals, texts, music pieces, etc. 
% synthesize
For the synthesis task, the input of the diffusion framework is the past human poses 
$x^{t}=\left\{m^{t-T_h}, m^{t+1-T_h}, \cdots ,m^t\right\}$, 
and the control input $c^{t}=\left\{a^{t-T_h}, a^{t+1-T_h}, \cdots ,a^{t+T_p}\right\}$, 
where $T_h$ denotes the length of the observed past poses and $T_p$ denotes the number of predicted frames. 
The output of the framework is the predicted future motion frames, written as $y^{t}=\left\{m^{t+1}, \cdots, m^{t+T_p}\right\}$. 
% reconstruct
For the reconstruction task, the past human poses are partially observed, e.g. with missing frames or missing body joints. In such cases, the task is to reconstruct a complete motion $x^t$ from an imperfect input, that we denote $\hat{x}^t$, with the control input $c^t$. 

\subsection{Motion Diffusion Models}
\label{sub:diffusion}

\SetKw{KwRepeat}{Repeat}
\SetKw{KwUntil}{Until}
\begin{algorithm}[t]
\caption{Training for data sample at time $t$}\label{alg:1}
\KwIn{data $y_0^t$, past motion $x^t$, and control input $c^t$}
\KwRepeat{}\\
  \quad  Initialize $s\sim$ Uniform($(1,\cdots,S)$, $\epsilon \sim \mathcal{N}(0, \mathbf{I})$\\
  \quad \textbf{if} diffusion dropout, $p\sim \mathcal{U}(0,1)$ \textbf{do}\\
  \quad \quad $x_d^t=x^t_s$ if $p<p_d$, else, $x_d^t=x^t$\\
  \quad  Take gradient step on \\
  \quad \quad  $\nabla_{\theta}\left\| \epsilon - \epsilon_{\theta}(\sqrt{\overline{\alpha}_s}y_0^t+\sqrt{1-\overline{\alpha}_s}\epsilon, s, x_d^t, c^t)  \right\|$ \\
\KwUntil{} converged
\end{algorithm}

\begin{figure*}[t]
\centering
\includegraphics[width=0.99\textwidth]{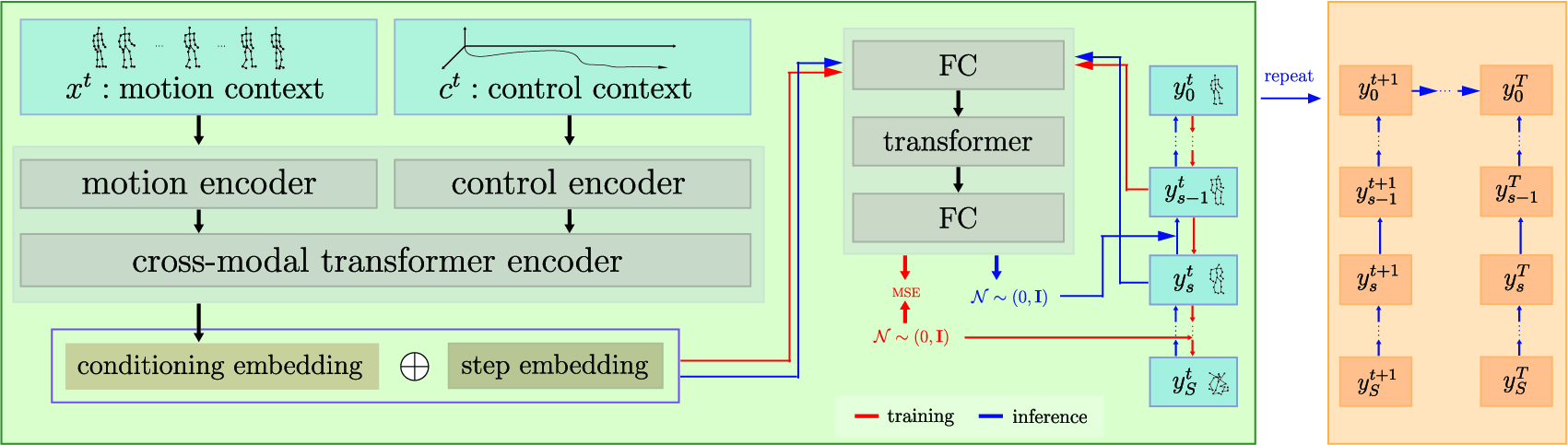}
\caption{Overview of the MoDiff schematic: a Transformer-based diffusion probabilistic model synthesizes future motion by the learned reverse diffusion process. The correlation between the motion and control context is depicted by the cross-modal transformer encoder. Future motions are predicted autoregressively. }
\label{fig:framework}
\end{figure*}

We address the formulated problem with our proposed autoregressive diffusion model (MoDiff) based on DDPM~\cite{ho2020denoising}, as illustrated in Fig.~\ref{fig:framework}. 
% Forward process
We define the forward diffusion process as $(y_0, y_1, \cdots, y_S)$, where $S$ is the maximum number of diffusion steps. For the sake of brevity, we omit the superscript $t$. 
The forward process is a stochastic process with a fixed Markov chain that gradually adds Gaussian noise to the ground truth future motion data 
$y_0 = y$ until the distribution of $y_S$ is close to a standard Gaussian distribution:
\begin{equation}
\begin{aligned}
    q(y_{1:S}|y_0) &:=  \prod_{s=1}^{S}q_{\theta}(y_{s}|y_{s-1});\\
    q(y_{s}|y_{s-1})&:=  \mathcal{N}(y_{s};\sqrt{1-\beta_{s}}y_{s-1}, \beta_{s}\mathbf{I}),
\end{aligned}
\end{equation}
where $\beta_{1},\beta_{2}, \cdots, \beta_{S}$ are the fixed variance schedulers for controlling the noise scale.
As shown in~\cite{ho2020denoising}, the forward diffusion sample at any diffusion step $s$ can be calculated in one step as: 
\begin{equation}
    q(y_{s}|y_{0}):= \mathcal{N}(y_{s};\sqrt{\overline{\alpha}_s}y_{0}, (1-\overline{\alpha}_s)\mathbf{I}),
    \label{equ:diffusion-process}
\end{equation}
where $\alpha_s=1-\beta_s$ and $\overline{\alpha}_s=\prod_{i=1}^{s}\alpha_i$.
% Reverse process
In the reverse generation process, we learn this process as $(y_S, y_{S-1}, \cdots, y_0)$ and generate motions by progressively denoising the pose from $y_S$ to $y_0$. 
We model this reverse generation process by parameterizing Gaussian transition probabilities with the past poses $x$ and the past and current control signal $c$ as conditioning information. The reverse generation process is formulated as:
\begin{equation}
\begin{aligned}
    p_{\theta}(y_{0:S}|x,c)&:=p(y_{S})\prod_{s=1}^{S}p_{\theta}(y_{s-1}|y_{s},x,c)\\
    p_{\theta}(y_{s-1}|y_{s},x,c)&:=\mathcal{N}(y_{s-1};\mu_{\theta}(y_{s},s,x,c),\Sigma_{\theta}(y_{s},s)),
\end{aligned}
\end{equation}
where $p(y_S)$ denotes a prior noise Gaussian distribution and $\Sigma_{\theta}(y_{s},s)=\beta_{s}\mathbf{I}$. 
All transitions share the same parameters. 

\subsection{Training and Inference}
\label{sub:strategy} 

\SetKw{KwFor}{for}
\begin{algorithm}[t]
\caption{Inference the future motions $y_0^t$}\label{alg:2}
\KwIn{noise $y_S^t\sim \mathcal{N}(0,\mathbf{I})$, past motion $x_t$, and control input $c^t$}
\KwFor{} $s=S,\cdots,1$, \textbf{do}\\
\quad   $y_{s-1}^t = \frac{1}{\sqrt{\alpha_s}}(y_s^t-\frac{\beta_s}{\sqrt{1-\overline{\alpha}_s}}\epsilon_{\theta}(y_s^t,s,x^t,c^t))+\sqrt{\beta_s}\mathbf{z}$\\
\quad   where $\mathbf{z} \sim \mathcal{N}(0,\mathbf{I})$ if $s>1$, else $\mathbf{z}=0$\\
\textbf{end for}\\
\KwOut{$y_0^t$}
\end{algorithm}

With the formulated forward diffusion process and reverse generation process, to generate the human pose of future motion frames $y_0$, the training process optimizes the log-likelihood in the reverse generation process by maximizing the variational lower bound:
\begin{equation}
\begin{aligned}
\mathbb{E}[\text{log}p_\theta(y_0)]&\ge \mathbb{E}_q[\text{log}\frac{p_\theta(y_{0:S}, x, c)}{q(y_{1:S}|y_0)}]\\
    &=\mathbb{E}_q[\text{log}p(y_S)+\sum_{s=1}^{S}\text{log}\frac{p_\theta(y_{s-1}|y_s, x, c)}{q(y_{s}|y_{s-1})}].
\end{aligned}
\end{equation}
We utilize the negative bound as the loss function,  written as the KL-divergence between Gaussian distributions: 
\begin{equation}
\begin{aligned}
\mathbb{E}_q[&-\text{log}p_\theta(y_0|y_1,x,c)+D_{KL}(q(y_S|y_0)||p(y_S))\\
&+\sum_{s=2}^{S}D_{KL}(q(y_{s-1}|y_s,y_0)||p_\theta(y_{s-1}|y_s,x,c))].
\end{aligned}
\end{equation}
The first KL-divergence term with $q(y_S|y_0)||p(y_S)$ has no learnable parameters and can thus be ignored. $q(y_{s-1}|y_s,y_0)$ is conditioning on $y_0$, which is tractable and can be represented as: 
\begin{equation}
\begin{aligned}
q(y_{s-1}|y_s, y_0)=
\mathcal{N}(y_{s-1};\tilde{\mu}_{s}(y_{s},y_0),\tilde{\beta}_{s}\mathbf{I}),
\end{aligned}
\end{equation}
where $\tilde{\mu}_{s}$ and $\tilde{\beta}_{s}$ is calculated as:
\begin{equation}
\begin{aligned}
\tilde{\mu}_{s}(y_{s},y_0)&=
\frac{\sqrt{\overline{\alpha}_{s-1}}\beta_s}{1-\overline{\alpha}_s}y_0+
\frac{\sqrt{\alpha_s}(1-\overline{\alpha}_{s-1})}{1-\overline{\alpha}_s}y_s\\
\tilde{\beta}_{s}&=\frac{1-\overline{\alpha}_{s-1}}{1-\overline{\alpha}_s}\beta_s\mathbf{I}.
\end{aligned}
\end{equation}
Ho et al. \cite{ho2020denoising} show that the second KL-divergence term can be calculated as:
\begin{equation}
\begin{aligned}
\mathbb{E}_q\left[ \frac{1}{2\beta_s}%{2\sigma_s^2} 
\left\| \tilde{\mu}_s(y_s,y_0)-\mu_\theta(y_s,s,x,c)\right\|^2\right]+\lambda,
\end{aligned}
\end{equation}
where $\lambda$ is a constant value that does not depend on $\theta$. 
We choose the reparameterization method shown in~\cite{ho2020denoising} that
\begin{equation}
\begin{aligned}
\mu_\theta(y_s, s, x, c)=\frac{1}{\sqrt{\alpha_s}}(y_s-\frac{\beta_s}{\sqrt{1-\overline{\alpha}_s}}\epsilon_\theta(y_s, s, x, c)),
\end{aligned}
\end{equation}
and the objective function is simplified to 
\begin{equation}
\begin{aligned}
\mathbb{E}_{\epsilon, y_0, s}\left\|\epsilon - \epsilon_{\theta}(\sqrt{\overline{\alpha}_s}y_0+\sqrt{1-\overline{\alpha}_s}\epsilon, s, x, c) \right\|^2,
\end{aligned}
\end{equation}
where $\epsilon_\theta$ is a network that predicts Gaussian noise $\epsilon\sim\mathcal{N}(\mathbf{0}, \mathbf{I})$ and is trained with the MSE loss. The training is performed for all steps $s\in[1,S]$. The complete training procedure with the simplified objective function is displayed as Algorithm~\ref{alg:1}, with the diffusion dropout later introduced in Section \ref{sub:diffusion dropout}. 
% Inference
After the network $\epsilon_\theta$ is trained, given past poses $x$, control input $c$, and $y_S \sim \mathcal{N}(\mathbf{0}, \mathbf{I})$, we can synthesize future motions $y_0$ 
% by sampling a Gaussian noise $y_s\sim\mathcal{N}(0,1)$ 
with the reverse generation process $y_{s-1}\sim p_\theta(y_{s-1}|y_s)$,:
\begin{equation}
\begin{aligned}
y_{s-1} = \frac{1}{\sqrt{\alpha_s}}(y_s-\frac{\beta_s}{\sqrt{1-\overline{\alpha}_s}}\epsilon_{\theta}(y_s,s,x,c))+\sqrt{\beta_s}\mathbf{z},
\end{aligned}
\end{equation}
where $\mathbf{z}\sim \mathcal{N}(\mathbf{0}, \mathbf{I})$ for $s\in[2,S]$ and $\mathbf{z}=\mathbf{0}$ when $s=1$. 
% autoregressive
We follow the generation procedure in Algorithm~\ref{alg:2} to predict the next sample $y_0^t=\{ m_0^{t+1},\cdots, m_0^{t+T_p} \}$. Inspired by Li et al.~\cite{li2021ai} and Guillermo et al.~\cite{valle2021transflower}, our model outputs the next $T_p$ poses given the past $T_h$ poses to improve model performance. 
We then only update the next time step $t+1$, pass it autoregressively to the transformer-based encoder together with the next control input, and repeat the same procedure until all desired motion frames are synthesized.

\subsection{Diffusion Data Dropout}
\label{sub:diffusion dropout}

Diffusion data dropout is applied during the training stage to improve data efficiency and model robustness. 
To be specific, we drop motion context information through the forward diffusion process without extra effort. 
A larger diffusion step $s$ in the diffusion data dropout leads to the past motion $x^t$ being more corrupted and less information is preserved. 
Similar to Equation \ref{equ:diffusion-process}, the past motion information with respect to step $s$ can be calculated in one step as:
\begin{equation}
    q(x_{s}^t|x^t):= \mathcal{N}(x_{s}^t;\sqrt{\overline{\alpha}_s}x^t, (1-\overline{\alpha}_s)\mathbf{I}). 
\end{equation}
As shown in Algorithm \ref{alg:1}, data dropout is done if $p < p_d$, where $p\sim\mathcal{U}(0,1)$ and $p_d$ is the diffusion dropout rate. 
We stabilize the training by starting with $p_d=0$, i.e., with complete motion context information. A diffusion dropout scheduler $P_d=\{p_{d1}, p_{d2},\cdots, p_{dn}\}$ is set to increase the dropout rate during training. As the diffusion dropout rate increases, the denoising process becomes more focused on the control context and more robust to corruption of the motion context.

\subsection{Motion Reconstruction}
For motion reconstruction, we use the same framework without any further training. Our framework allows an imperfect input $\hat{x}^t=\{\hat{m}^{t-T_h},\cdots,\hat{m}^{t}\}$ since diffusion data dropout was applied during training.
To reconstruct missing body joints or frames, we first generate a sequence of $T_h$ future frames, $\{{m}^{t+1},\cdots,{m}^{t+T_h}\}$. %future sequences.
Then we reverse the order of this sequence and the control signals  $\{a^{t-T_h},\cdots,a^{t+T_h}\}$ to $\{{m}^{t+T_h},\cdots,{m}^{t+1}\}$ and $\{a^{t+T_h},\cdots,a^{t-T_h}\}$, 
exploiting the fact that the training data are augmented by lateral mirroring and time-reversion. 
We regard the reversed motion context and control context as conditioning information to generate poses. The missing information can be reconstructed by filling the holes with the generated parts. We repeat until all missing parts are reconstructed. 

\subsection{Network Architecture}
\label{sub:architecture}

The proposed MoDiff framework is composed of an encoder and a decoder. 
The encoder encodes the past motion context and control context, and the decoder models the Gaussian transitions in the Markov chain, as depicted in Fig.~\ref{fig:framework}. 
For the encoder, we design two-layer transformers that encode the motion context $x^t$ and control context $c^t$ separately, with output embeddings of the same dimension. The motion context is augmented by diffusion dropout. In the transformers, a position embedding is introduced to emphasize the positional relation at different time steps. The outputs are then concatenated together into a six-layers cross-modal transformer. 
For the decoder, we design a three-layer Transformer decoder similar to \cite{gu2022stochastic} to model spatial and temporal dependencies. The inputs include the future motion $y^t$, the noise variable $\epsilon$, a diffusion step index $s$, as well as the output embedding of the cross-model Transformer. In diffusion step $s$, the noised future motion $y_s^t$ is concatenated with a diffusion step embedding. Finally, fully-connected layers downsample the output to the motion dimension. Like in TimeGrad~\cite{rasul2021autoregressive}, we can pass autoregressively to the network to repeat until the desired horizon although TimeGrad primarily relies on recurrent nets for propagating temporal information.  

\section{EXPERIMENTS AND EVALUATIONS}
\label{sec:exp}

In this section, We first describe the experimental setting, including the dataset, ablation settings, and implementation details. We then present  the results on the generation and reconstruction of human locomotion samples. 
On the basis of these, we evaluate and discuss the performance of both the proposed frameworks and the baselines, followed by preliminary results on various tasks.

\begin{figure}[b]
\centering
\includegraphics[width=0.48\textwidth]{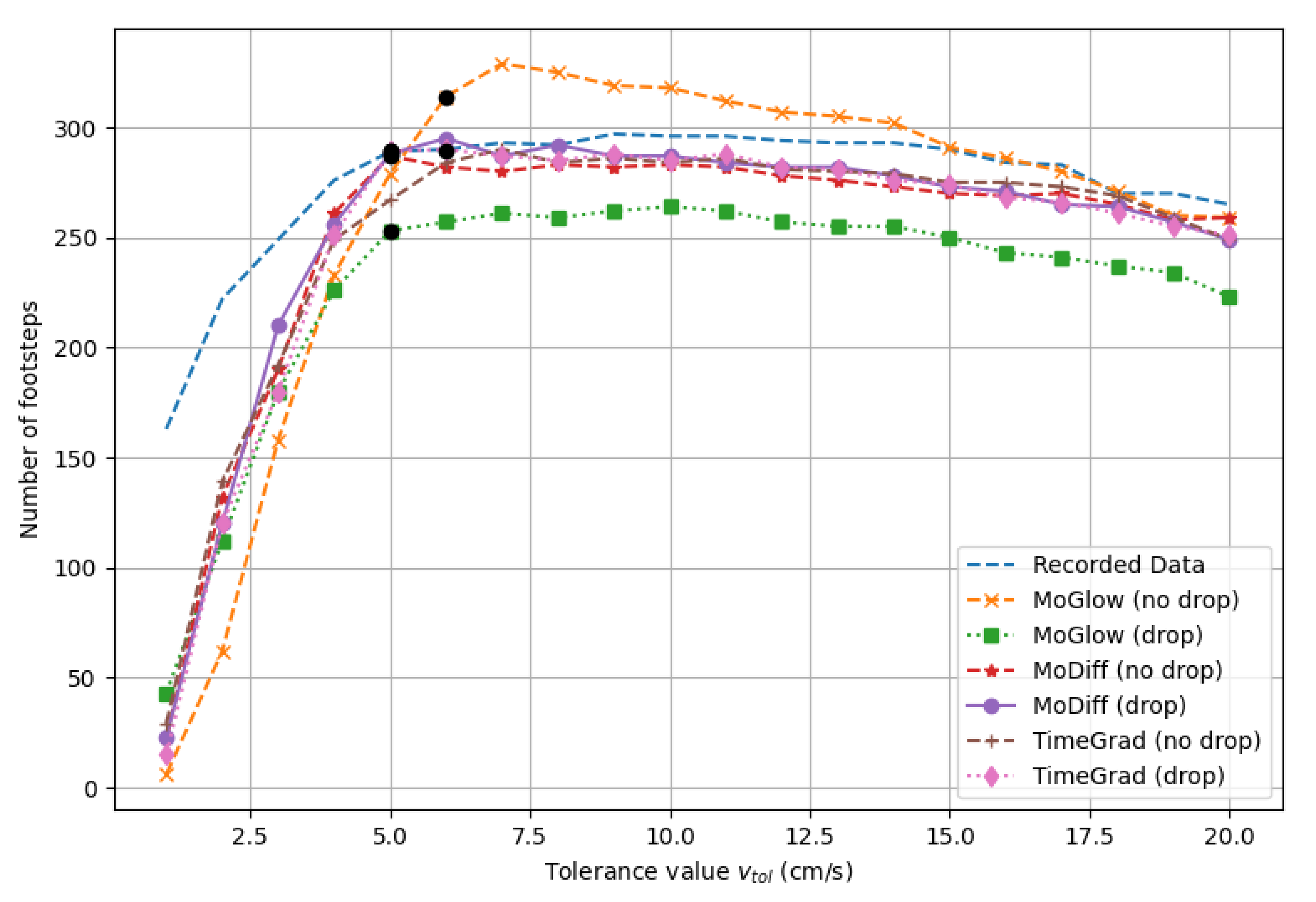}
\caption{ 
    Footstep analysis for generated motion given past information and control signals. The footstep count $f_{est}$ on tolerance value $v_{tol}$. The black dots indicate $v_{tol}^{95}$, the first velocity tolerance for capturing 95\% estimated footsteps.
  }
\label{fig:fta}
\end{figure}

\begin{figure}[b]
\centering
\includegraphics[width=0.48\textwidth]{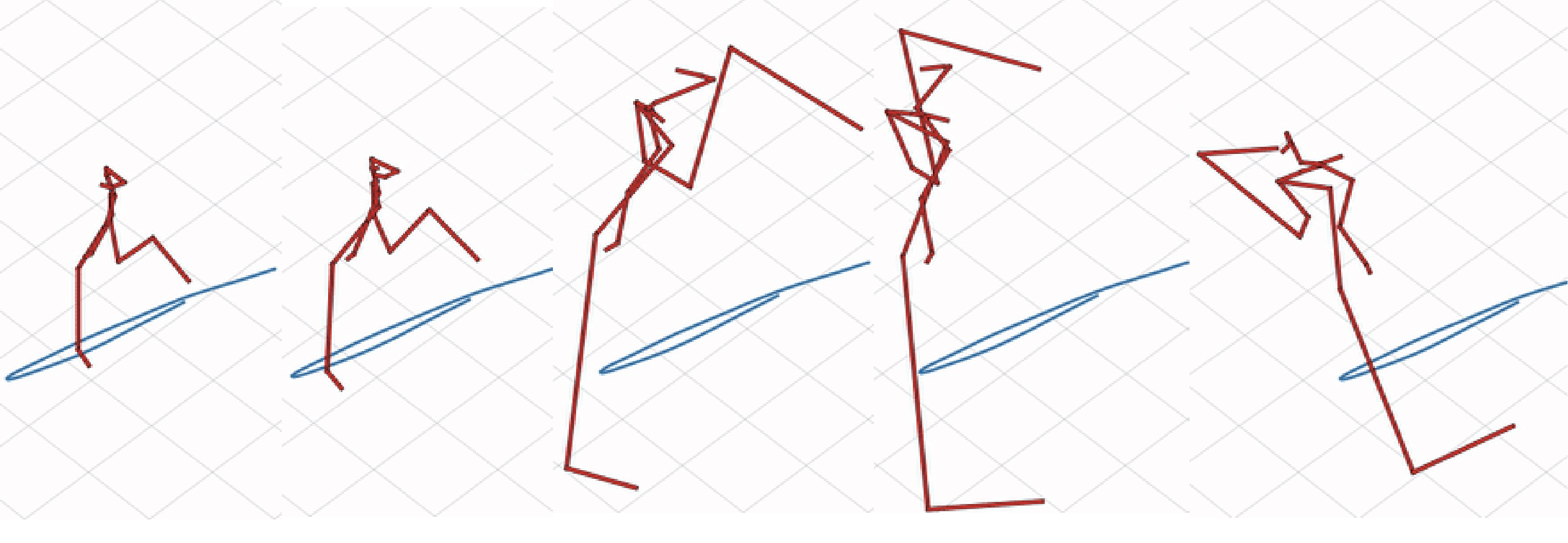}
\caption{ 
    Example poses with body joints flying away that are generated by MoGlow with incomplete past poses.  
  }
\label{fig:fly}
\end{figure}

\subsection{Experimental Setting}
\textbf{Dataset.}
We evaluate MoDiff on a dataset of human locomotion preprocessed by~\cite{henter2020moglow} that retargets skeletons from the Edinburgh Locomotion MoCap, CMU Motion Capture, and HDM05 datasets, which includes various human gaits along different trajectories. The motion context $m\in \mathbb{R}^{63}$ is represented by the 3D coordinates of 21 body joints. The control context $a\in \mathbb{R}^3$ includes forward, lateral and rotational velocities. We slice the training data into 4-second clips and downsample these to 20 fps with 50\% overlap. For synthesis and reconstruction with incomplete input, we generate data with missing parts by setting some joints to zero. 

\textbf{Ablation Settings.}
To assess the impact of design decisions, we compare our proposed MoDiff with MoGlow~\cite{henter2020moglow} and TimeGrad~\cite{rasul2021autoregressive}. 
MoGlow is an autoregressive model based on normalizing flows and LSTM for human motion generation. 
TimeGrad is an autoregressive diffusion model for general time series forecasting and reports state-of-the-art performance on real-world datasets, such as traffic and electricity. %~\cite{salinas2019high}. 
Both MoDiff and MoGlow are probabilistic and designed for controllable motion synthesis. 
Unlike MoGlow, MoDiff and TimeGrad are diffusion-based models, but the encoder of TimeGrad is still based on LSTM, while MoDiff uses transformers. 
We evaluate the impact of the diffusion models and transformers architecture by comparing MoDiff with MoGlow and TimeGrad. 
To highlight the advantage of the proposed diffusion data dropout, we applied this dropout strategy to both MoDiff and the baseline models.

\textbf{Implementation Details.}
Following the experimental setting in MoGlow~\cite{henter2020moglow}, MoDiff is trained with 10-frame time windows, i.e., $T_h=10$. 
In our experiments, the maximum number of diffusion steps $S$ is 100. 
The diffusion dropout rate is $\{0,0.05,\cdots,0.25\}$, increasing every 100 epochs after 500 epochs. 
The Transformer in our experiments is set to 256 dimensions and 4 heads.
We use the standard Transformer encoder in PyTorch, with the T5-style relative positional embedding.
We train with Adam optimizer with a learning rate of 5e-5 and batch size of 64 for 1000 epochs. All experiments were conducted on a single NVIDIA A100 Tensor Core GPU.

\begin{table}[t]
\caption{Results of the footsteps and bone-length analysis with complete input. The number closest to the recorded motion capture is in bold.\\}
\resizebox{0.99\linewidth}{!}{
\begin{tabular}{l|ccccc}
\hline
Model                  & $f_{est}$    & $v_{tol}^{95}$ & $\mu$          & $\sigma$       & RMSE           \\ \hline
Recorded Data                 & 289          & 5              & 0.315          & 0.263          & 0              \\ \hline
MoGlow (no drop)       & 314          & 6              & 0.222          & 0.134          & 0.315          \\
MoGlow (drop)          & 253          & \textbf{5}    & 0.341          & 0.250          & 0.250          \\
TimeGrad (no drop)  & \textbf{284}   & 6    & \textbf{0.331 }    & 0.338       & 0.124        \\
TimeGrad (drop)     & 281   & \textbf{5}     & 0.289     & 0.241       & 0.115         \\
MoDiff (no drop)       & 280          & \textbf{5}    & 0.289          & 0.239       & 0.089          \\
MoDiff (drop) & \textbf{284} & \textbf{5}     & 0.334 & \textbf{0.286} & \textbf{0.072 }\\ \hline
\end{tabular}}
\label{tab:1}

\end{table}

\begin{table}[t]
\caption{Footsteps and bone-length results of generation and reconstruction with incomplete input. The number closest to the recorded motion capture is in bold.}
\resizebox{0.99\linewidth}{!}{
\begin{tabular}{l|cccccc}
\hline
Model                  & $f_{est}$    & $v_{tol}^{95}$ & $\mu$          & $\sigma$       & RMSE$_{ge}$    & RMSE$_{re}$    \\ \hline
Recorded Data                 & 289          & 5              & 0.315          & 0.263          & -              & -              \\ \hline
MoDiff (no drop)       & 299          & \textbf{5}     & 0.375          & 0.347          & 0.105          & 0.116          \\
\textbf{MoDiff (drop)} & \textbf{293} & \textbf{5}     & \textbf{0.358} & \textbf{0.287} & \textbf{0.101} & \textbf{0.108} \\ \hline
\end{tabular}}
\label{tab:2}
\end{table}

\begin{figure*}[t]
\centering
\includegraphics[width=0.99\textwidth]{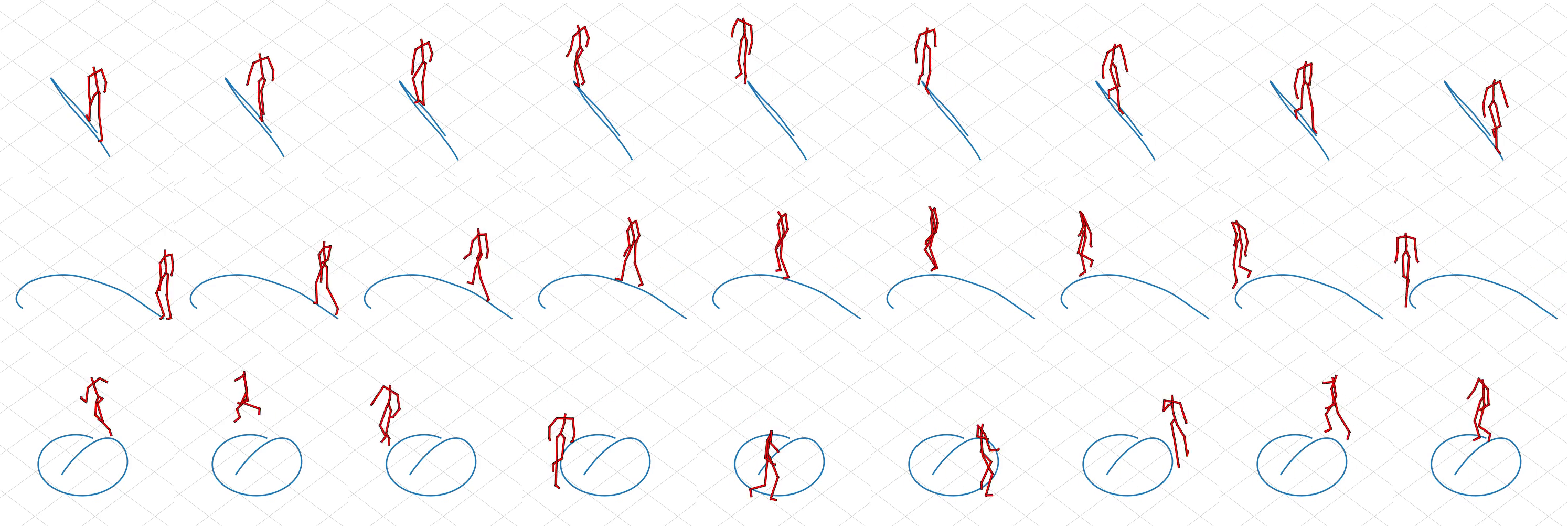}
\caption{Example sequences generated by MoDiff given different past information and control signals. }
\label{fig:div}
\end{figure*}

\begin{figure*}[t]
\centering
\includegraphics[width=0.99\textwidth]{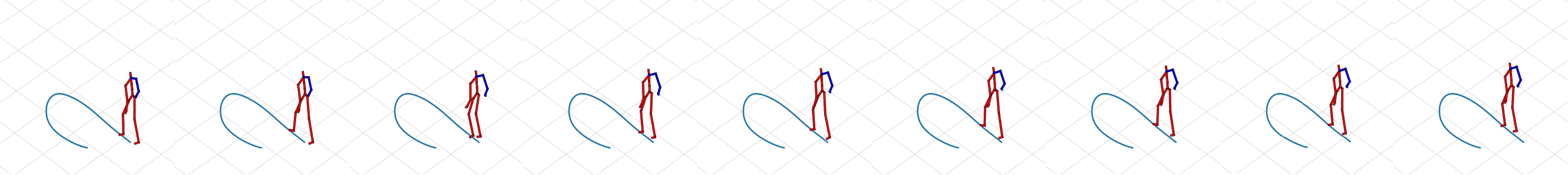}
\caption{An example sequence reconstructed by MoDiff. The right arm is missing in the past frames. }
\label{fig:reco}
\end{figure*}

\subsection{Results and Discussions}
\textbf{Results.}
For quantitative analysis, we evaluate the quality of motion generation with footstep analysis and bone-length analysis, which are widely-used methods to evaluate artifacts related to heel sliding and limb stretching~\cite{henter2020moglow}. 
Footstep analysis compares the estimated footsteps of the generated motion and the recorded motion. 
The average number of footsteps $f_{est}$ can be detected at intervals where the horizontal velocity of the heel joints is below a tolerance value $v_{tol}$. For evaluation, we compare the estimated number of footsteps at the first velocity tolerance for capturing 95\% steps, denoted as $v_{tol}^{95}$. The estimated number of footsteps on different $v_{tol}$ in the generated motions are displayed in Fig.~\ref{fig:fta}, the detected $v_{tol}^{95}$ is shown as black dot in this figure. We can observe that the curves of MoDiff and TimeGrad, with and without diffusion dropout, are consistently closer to the curve of the recorded motion compared to MoGlow.
Table~\ref{tab:1} shows the results given complete context, include the estimated footsteps$f_{est}$, the tolerance value $v_{tol}^{95}$, the mean $\mu$ and standard deviation $\sigma$ of the step duration, as well as the results of bone-length analysis. 
The bone-length analysis is for evaluating bone-stretching artifacts. We compare the RMSE ($cm$) of the bone length. 
Table~\ref{tab:2} shows the results of generation and reconstruction given incomplete context.

\textbf{Discussion.}
% Qualitatively and quantitatively, on the locomotion task, MoDiff can generate recorded realistic motion, and have good diversity, as shown in Fig.~\ref{fig:div}.
MoDiff can qualitatively and quantitatively generate recorded realistic motion with good diversity on the locomotion task, as demonstrated by the snapshots of generated motion sequences in Fig.~\ref{fig:div}. 
In contrast, the native MoGlow suffers from poor adherence and generates motion with noticeable foot-sliding artifacts. %Moreover, MoDiff outperforms TimeGrad on the RMSE metric, while being on par on other metrics. 
As can be seen from the results in Table~\ref{tab:1}, the diffusion-based models, TimeGrad and MoDiff, perform significantly better in terms of both footstep and bone-length analysis compared to MoGlow which is based on normalizing flow. However, unlike the models the use LSTM, i.e.~MoGlow and TimeGrad, MoDiff benefits from transformer-based encoders to further improve on bone-length analysis.
%The major difference in modeling is that MoGlow relies on normalizing flow and LSTM, and TimeGrad relies on diffusion-based models and LSTM, whereas our MoDiff architecture benefits from using diffusion models and transformers. 
%This is further supported by the results presented in Table~\ref{tab:1}, which highlight the necessity of using transformer-based encoders against TimeGrad. 
The attention mechanisms in MoDiff discern specific features over a longer period, which is essential for learning consistent gaits under conditioning signals. Additionally, the attention-based transformer is effective at extracting meaningful information from intermediate diffusion steps with roughly corrected skeleton poses, which is not as feasible with standard LSTM models.

MoDiff can be extended to reconstruct incomplete body joints or frames without extra training, as illustrated by the reconstruction results in Table~\ref{tab:2}, results that do not significant diverge from those in Table~\ref{tab:1}. When the past input is incomplete, MoGlow exhibits unstable outputs, i.e. skeletons with joints flying away, as shown in Fig.~\ref{fig:fly}. The results from MoDiff are closer to the recorded motion capture, illustrating the robustness of the diffusion-based architecture. From the example shown in Fig.~\ref{fig:reco}, we observe that the reconstructed body joints (the right arm in blue) fit the original skeleton well. 
As can be seen from both Table~\ref{tab:1} and Table~\ref{tab:2}, the proposed data dropout strategy improves the performance of all three methods, which confirms its effectiveness.
%Summarising Table~\ref{tab:1} and Table~\ref{tab:2}, the proposed data dropout method moderately improves the performance of MoDiff, as well as the native MoGlow and TimeGrad, which confirms its effectiveness. %We conclude further improvement from the contributed diffusion dropout in comparison with the original methods, showing an extra advantage by leveraging the diffusion-based mixture representations. 

\textbf{Extra Applications.}
MoDiff is a task-agnostic framework that can be applied to other cross-modal generation tasks, such as text-to-motion and music-to-dance. We show generated samples in the attached multimedia materials link on \href{https://youtu.be/qeAs9eF3pbs}{YouTube:youtu.be/qeAs9eF3pbs}.
Note that the framework was not intended to be text-based even if it is agnostic to the input modality. The text here is just to generate a command token that can be dynamic and from other control modalities, e.g., keyboard strokes. It is thus more appropriate to compare MoDiff with other works for controllable locomotion generation. 

\section{CONCLUSIONS}
\label{sec:conclu}

We propose MoDiff, a Transformer-based diffusion model, to tackle the challenge of controllable and robust human motion synthesis and reconstruction under imperfect conditions. 
We introduce a novel diffusion data dropout strategy utilizing the diffusion forward process, which improves data efficiency and model robustness. The comparison results on the locomotion dataset with state-of-the-art baselines demonstrate the superiority of our MoDiff.
MoDiff can be applied to various multimodal synthesis tasks. 
In the future, we plan to extend the MoDiff framework for classifier-guided conditional generation and apply it for more challenging dance motions.

\section*{ACKNOWLEDGMENT}

This study has received funding from the European Commission Horizon 2020 research and innovation program under grant agreement number 824160 (EnTimeMent). This work benefited from access to the HPC resources provided by the Swedish National Infrastructure for Computing (SNIC), partially funded by the Swedish Research Council through grant agreement no. 2018-05973.

\bibliographystyle{IEEEtran}

\bibliography{IEEEexample}

\end{document}